%% file: main.tex
\documentclass[conference]{IEEEtran}

\usepackage{times}

\usepackage[numbers]{natbib}
\usepackage{multicol}
\usepackage[bookmarks=true, allcolors=blue]{hyperref}

\let\labelindent\relax
\usepackage{enumitem}
\usepackage{subfiles}
\usepackage{graphicx}
\usepackage{caption}
\usepackage{bm}
\usepackage{xcolor}
\usepackage{xspace}
\usepackage{amsmath}
\usepackage{amssymb}
\usepackage{multirow}
\usepackage{algorithm}
\usepackage{algorithmicx}
\usepackage[free-standing-units=true]{siunitx}
\usepackage[noend]{algpseudocode}

\ifCLASSOPTIONcompsoc
  \usepackage[caption=false,font=normalsize,labelfont=sf,textfont=sf]{subfig}
\else
  \usepackage[caption=false,font=footnotesize]{subfig}
\fi
\usepackage{balance}

\newcommand{\comment}[1]{}

\algnewcommand\algorithmicswitch{\textbf{switch}}
\algnewcommand\algorithmiccase{\textbf{case}}
\algdef{SE}[SWITCH]{Switch}{EndSwitch}[1]{\algorithmicswitch\ #1\ \algorithmicdo}{\algorithmicend\ \algorithmicswitch}%
\algdef{SE}[CASE]{Case}{EndCase}[1]{\algorithmiccase\ #1}{\algorithmicend\ \algorithmiccase}%
\algtext*{EndSwitch}%
\algtext*{EndCase}%

\makeatletter
\DeclareRobustCommand\onedot{\futurelet\@let@token\@onedot}
\def\@onedot{\ifx\@let@token.\else.\null\fi\xspace}

\makeatother

\begin{document}

\title{Augmenting differentiable physics \\ with randomized smoothing}

\author{\authorblockN{Quentin Le Lidec$^{1}$,
  Louis Montaut$^{1,2}$,
  Cordelia Schmid$^{1}$,
  Ivan Laptev$^{1}$,
  Justin Carpentier$^{1}$}
\authorblockA{$^1$ Inria and Département d'Informatique de l'\'Ecole Normale Supérieure,\\
PSL Research University, Paris, France\\
{\tt\small firstname.lastname@inria.fr}
}

\authorblockA{$^2$ Czech Institute of Informatics, Robotics and Cybernetics,\\
Czech Technical University, Prague, Czech Republic\\
{\tt\small firstname.lastname@cvut.cz}
}
}
\maketitle

\begin{abstract}
  \subfile{sections/0_abstract/abstract.tex}
\end{abstract}

\IEEEpeerreviewmaketitle

% SECTION I
\section{Introduction}
\label{sec:introduction}

\subfile{sections/1_introduction/intro.tex}
% --> END SECTION I

% SECTION III
\section{Differentiable rendering}
\label{sec:diff_render}

\subfile{sections/3/diff_rendering.tex}

% --> END SECTION III

% SECTION IV
\section{Differentiable simulation for control}
\label{sec:diff_sim}

\subfile{sections/4/diff_sim.tex}
% --> END SECTION IV

% CONCLUSION
\section{Discussion}
\label{sec:discussion}
\subfile{sections/5_discussion/discussion.tex}

\section*{Acknowledgments}
{
This work was supported in part by the French government under management of Agence Nationale de la Recherche as part of the "Investissements d'avenir" program, reference ANR-19-P3IA-0001 (PRAIRIE 3IA Institute), the European Regional Development Fund under the project IMPACT (reg. no. CZ.02.1.01/0.0/0.0/15 003/0000468), the Grant Agency of the Czech Technical University in Prague, grant No. SGS21/178/OHK3/3T/17, and Louis Vuitton ENS Chair on Artificial Intelligence.
}
%% Use plainnat to work nicely with natbib. 

%\newpage
\balance{}
\bibliographystyle{plainnat}
{\footnotesize
\bibliography{references}
}

\end{document}

%% file: sections/0_abstract/abstract.tex
In the past few years, following the differentiable programming paradigm, there has been a growing interest in computing the gradient information of physical processes (e.g., physical simulation, image rendering).
However, such processes may be non-differentiable or yield uninformative gradients (i.d., null almost everywhere).
When faced with the former pitfalls, 
gradients estimated via analytical expression or numerical techniques such as automatic differentiation and finite differences, make classical optimization schemes converge towards poor quality solutions.
Thus, relying only on the local information provided by these gradients is often not sufficient to solve advanced optimization problems involving such physical processes, notably when they are subject to non-smoothness and non-convexity issues.
In this work, inspired by the field of zero-th order optimization, we leverage 
randomized smoothing to augment differentiable physics by estimating gradients in a neighborhood. 
Our experiments suggest that integrating this approach inside optimization algorithms may be fruitful for tasks as varied as mesh reconstruction from images or optimal control of robotic systems subject to contact and friction issues.

%% file: sections/1_introduction/intro.tex
\subsection{Motivations}
Various physical processes such as image rendering \cite{opendr,liu2019softras} or physical simulation \cite{todorov2012mujoco, bullet}, are subject to non-smooth phenomena. 
These result in badly conditioned gradients which may be uninformative (i.d., null almost everywhere) or even undefined. 
Typically, rendering \cite{opendr,liu2019softras} is naturally non-differentiable and yields either null or undefined derivatives.
Similarly, differentiable simulators exhibits null gradients over entire regions of the trajectory space \cite{werling2021fast}.
In both cases, these physical gradients present some discontinuities or lack of regularity, which, in the end, may drastically hinder gradient-based optimization algorithms. 
In this presentation, we show how randomized smoothing appears as a natural solution to avoid the pitfalls induced by a lack of regularity originating from classic gradient estimators.
We notably show its applications in the context of differentiable rendering~\cite{le2021diffrender} and differentiable simulation for control~\cite{lidec2022leveraging}.

\subsection{Background on randomized smoothing}
Originally used in black-box optimization algorithms~\cite{matyas1965random,polyak1987randomopt}, randomized smoothing was recently introduced in the machine learning community \cite{duchi2012randomized,berthet2020learning} in order to integrate non-differentiable operations inside neural networks.

In more details, a function $g$ can be approximated by convolving it with a probability distribution $\mu$:
\begin{equation}
    g_\epsilon(x) = \mathbb{E}_{Z\sim \mu} \left[ g(x + \epsilon Z) \right]
\end{equation}
which corresponds to the randomly smoothed counterpart of $g$ and can be estimated with a Monte-Carlo estimator as follows:
\begin{equation}
    g_\epsilon(x) \approx \frac{1}{M} \sum_{i=0}^M g(x + \epsilon z^{(i)})
\end{equation}
where $\{ z^{(1)}, \dots, z^{(M)}\}$ are i.i.d. samples and $M$ is the number of samples.
Intuitively, the convolution makes $g_\epsilon$ smoother than its original counterpart $g$ and, thus, yields better conditioned gradients. 

Using an integration by part yields a zero-th order estimator:
\begin{equation}
        %\nabla_x g_\epsilon(x) &= \mathbb{E}_{Z \sim \mu} \left[ -g(x + \epsilon Z) \frac{\nabla \log \mu (Z)^\top}{\epsilon} \right], \label{eq:zero_RS}\\
     {\nabla^{(0)}_x \, g_\epsilon(x)} =  \frac{1}{M} \sum_{i=0}^M -g(x + \epsilon z^{(i)}) \frac{\nabla \log \mu (z^{(i)})^\top}{\epsilon}. \label{eq:zero_RS_est}
\end{equation}
This zero-th order estimator can be used even when~$g$ is non differentiable to obtain first-order information usable in a gradient-based optimization scheme, as in Sec.~\ref{sec:diff_render}.

Alternatively, a direct way to estimate the gradients whenever $g$ is differentiable, is to use the first-order estimator:
\begin{equation}
        %\nabla_x g_\epsilon(x) &= \mathbb{E}_{Z\sim \mu} \left[ \nabla g(x + \epsilon Z) \right], \label{eq:first_RS}\\
     {\nabla^{(1)}_x \, g_\epsilon(x)} =  \frac{1}{M} \sum_{i=0}^M \nabla g(x + \epsilon z^{(i)}). \label{eq:first_RS_est}
\end{equation}
Such a first-order estimator averages gradients over a noise distribution and, thus, mechanically augments the quantity of captured information in the gradients of~$g$ even when they are uninformative e.g null in a region of the optimization space, as in Sec.~\ref{sec:diff_sim}. 

%% file: sections/3/diff_rendering.tex
\begin{figure}
\includegraphics[width=0.8\linewidth]{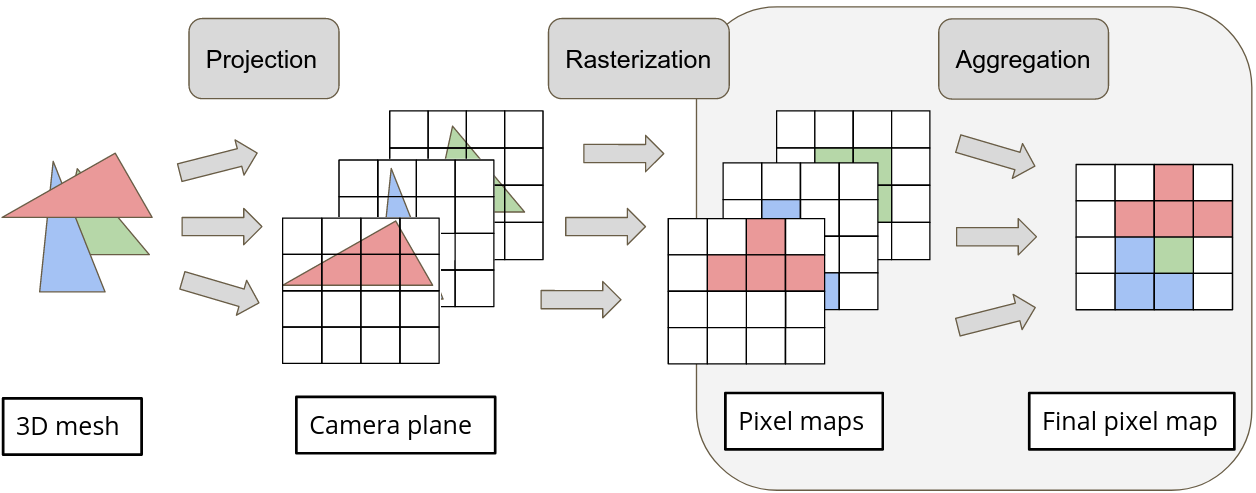}
\includegraphics[width=0.8\linewidth]{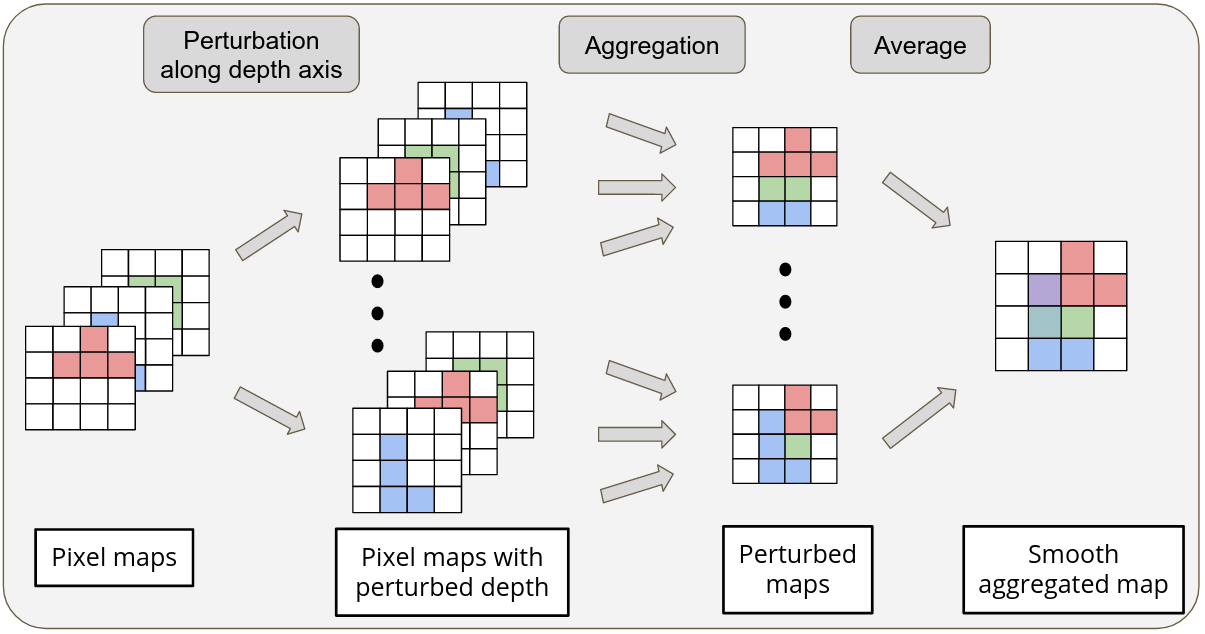}
\centering
\caption{\textbf{Top}: Overview of the rendering process: both rasterization and aggregation steps induce non-smoothness in the computational flow. \textbf{Bottom:} Illustration of the differentiable perturbed aggregation process. The rasterization step is made differentiable in a similar way.}
\label{fig:rendering_steps}
\vspace{-0.5cm}
\end{figure}

An example of a non-differentiable process which can be turned into a differentiable one via zero-th order randomized smoothing is rendering.
Rendering is the process which, given a 3D scene and a camera parameterization, outputs the corresponding RGB image.
As illustrated in Fig.~\ref{fig:rendering_steps}, rasterization-based renderers need to go through two discontinuous steps, called rasterization and aggregation.
These discontinuities seem natural: if an object moves on a plane parallel to the camera, some pixels will immediately change color at the moment the object enters the camera view or becomes unocluded by another object.
Mathematically, the operations occurring during these steps can be written as \textit{argmin} operators of Linear Programming problems:
\begin{align}
    y^*(\theta) = \underset{y\in \mathcal{C}}{\text{argmin}}\ \langle \theta, y \rangle
\end{align}
Such operators have null gradients almost everywhere and undefined otherwise, hence making them unusable for optimization.

Differentiable rendering approximates the underlying process by a smoother one by adding blur and transparency effects \cite{liu2019softras,le2021diffrender,gradsim}.
In particular, in \cite{le2021diffrender}, we propose to do it by replacing the original $y^*(\theta)$ operators from rasterization and aggregation by their randomly smoothed counterparts (Fig.~\ref{fig:rendering_steps}).
If randomized smoothing alleviates the burden of badly conditioned gradients by working with an approximate but smoother renderer, it is necessary to consider a way to reduce the smoothing during optimization in order to solve the original problem.
In \cite{le2021diffrender}, we introduce a criterion allowing to automatically reduce the injected noise across iterations. 
This approach is applied to tackle pose optimization and mesh reconstruction tasks and reaches results competitive with the state of the art \cite{le2021diffrender}.

%% file: sections/4/diff_sim.tex
\begin{figure}[t]
    \centering
    \includegraphics[width=0.95 \linewidth]{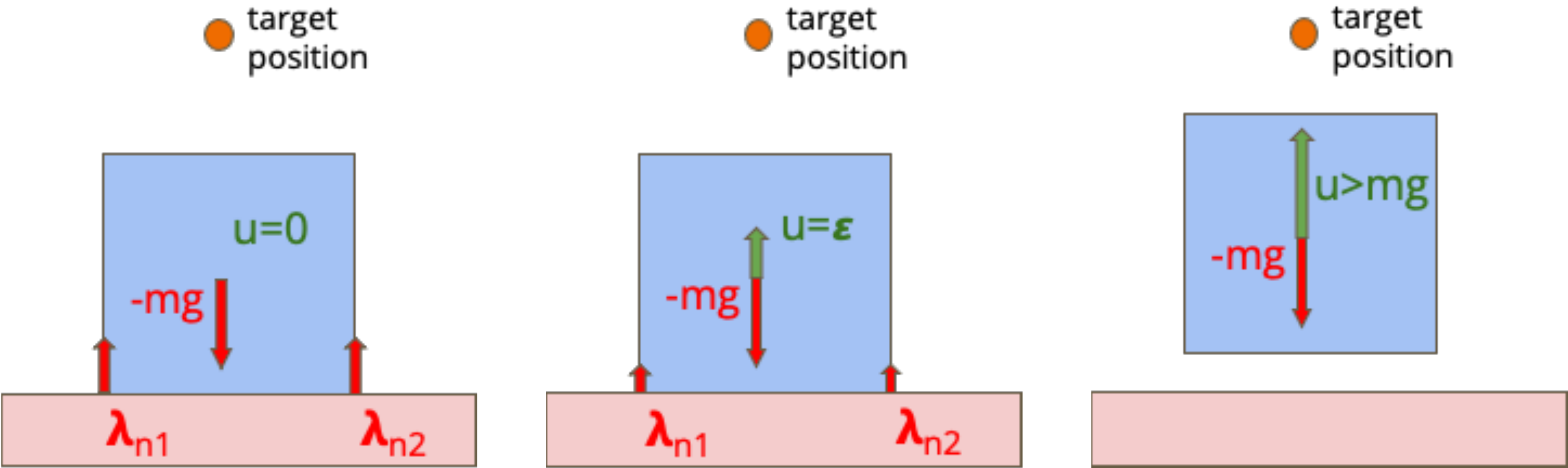}\\
    \includegraphics[width=0.8 \linewidth]{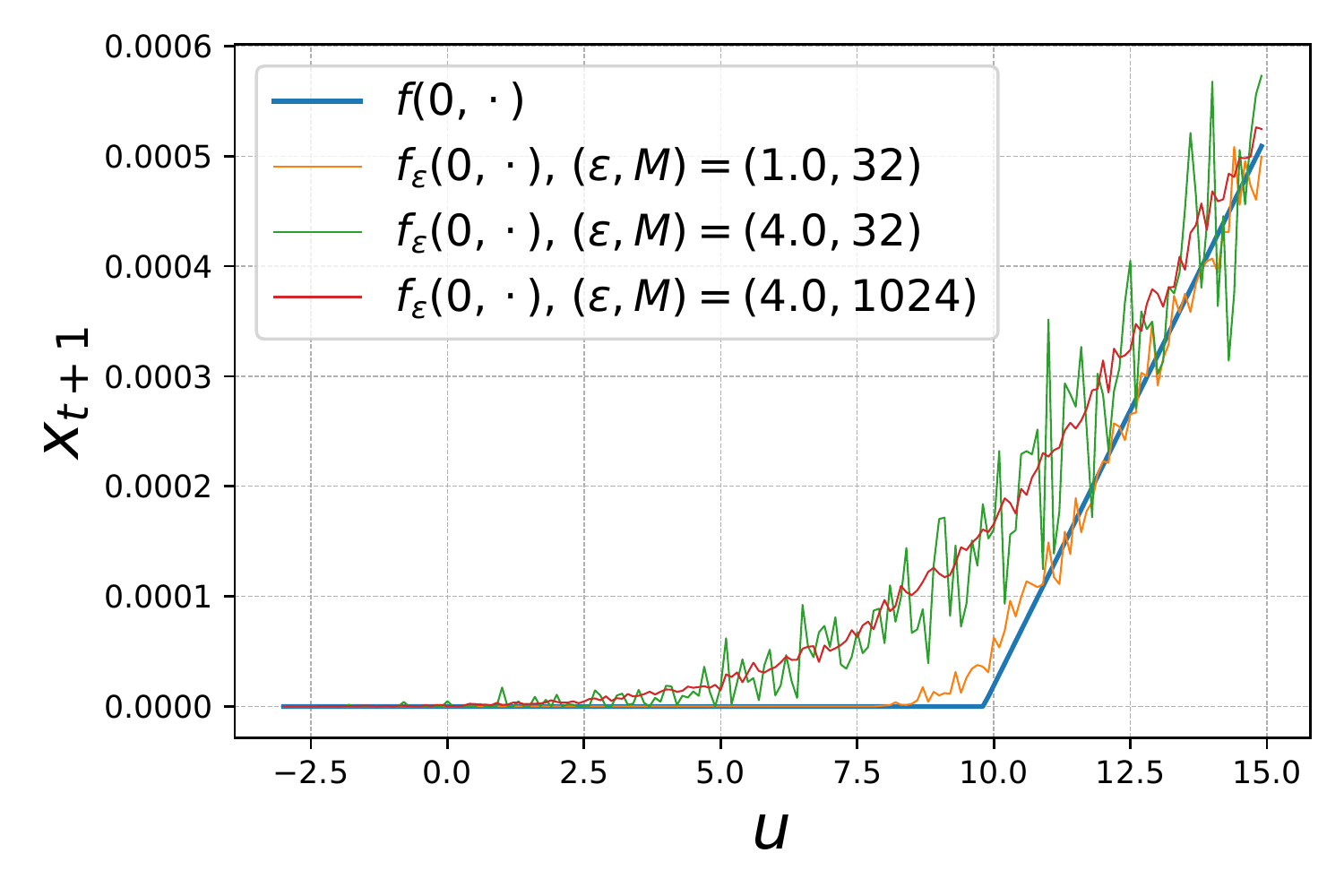}
    \caption{\textbf{Top:} A slight vertical force is not able to break the unilateral contact, leaving cube on the floor and, thus, the state unchanged. \textbf{Bottom:} The non-smoothness of physics induces null gradients %$\nabla_{\boldsymbol{u}} R$
    which results in the failure of classical optimization techniques} %(\ref{sec:local_OC}).}
    \label{fig:cube_schema}
    \vspace{-0.5cm}
\end{figure}
An example of a differentiable process for which first-order gradient estimation via randomized smoothing can be applied to enrich gradients' information is physical simulation.
Simulating a physical system subject to contacts and frictions requires to solve a problem formed by the Euler-Lagrange equations of motion, the complementarity constraint accounting for contacts, the Maximum Dissipation Principle and Coulomb's law which governs friction.
Solutions of this problem are typically obtained by solving an associated optimization problem \cite{todorov2012mujoco, de2018end, le2021diffsim} and, then, their gradients can be computed via implicit differentiation.
However, as illustrated by Fig.~\ref{fig:cube_schema} and detailed in \cite{werling2021fast}, contacts and friction cause the simulation to be non-smooth and its gradients to remain null over large regions of the trajectory space.

In such a setting, randomized smoothing allows to approximate the underlying physics by a smoother one as shown in \cite{lidec2022leveraging,suh2022bundled} and illustrated by Fig.~\ref{fig:cube_schema}.
Interestingly, this approach also allows to frame the stochasticity of exploration as a key element of the success of RL on  solving physical problems with contacts and frictions \cite{lidec2022leveraging}.
If randomized smoothing allows to augment the quantity of information contained in the gradient estimator, it also introduces stochasticity and it is necessary to design accordingly an optimization procedure that preserve the convergence guarantees.
To enforce the convergence towards an optimal (local) solution, it remains crucial to reduce the noise injected via the randomized smoothing across the iterations. 
In \cite{lidec2022leveraging}, we propose to decrease it in a way that adapts to the problem and avoids the smoothing being reduced too quickly, which would lead to performances similar to classical DDP, or too slowly, which would induce an unnecessary large number of iterations.

\begin{figure*}[t]
    \centering
    \includegraphics[width=0.15 \textwidth]{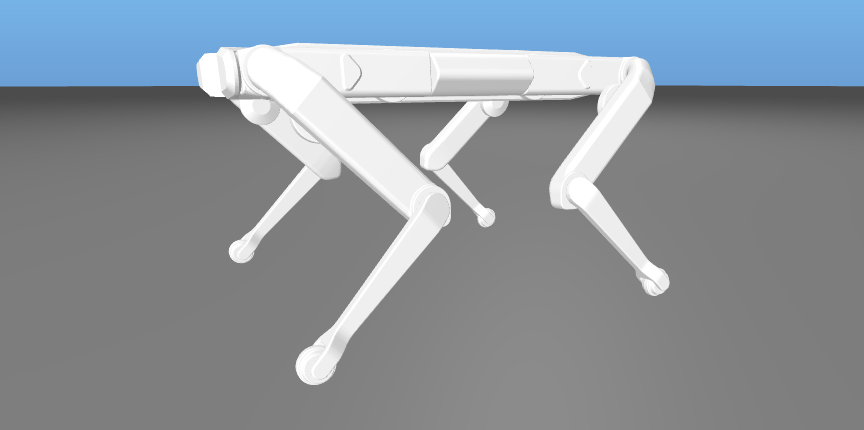}
    \includegraphics[width=0.15 \textwidth]{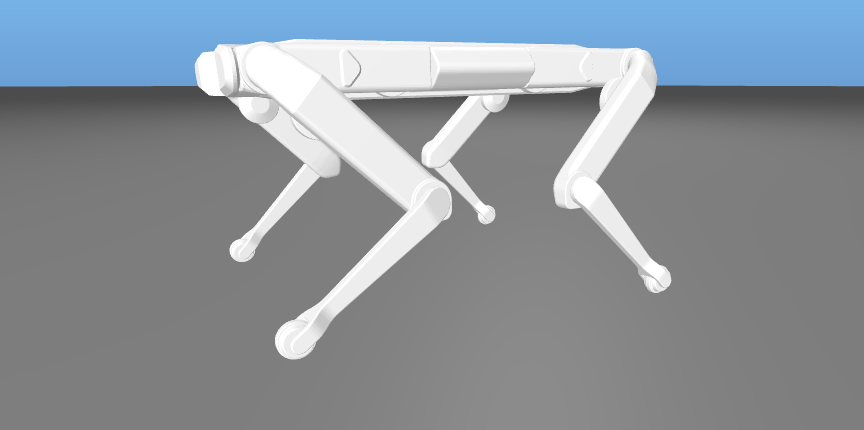}
    \includegraphics[width=0.15 \textwidth]{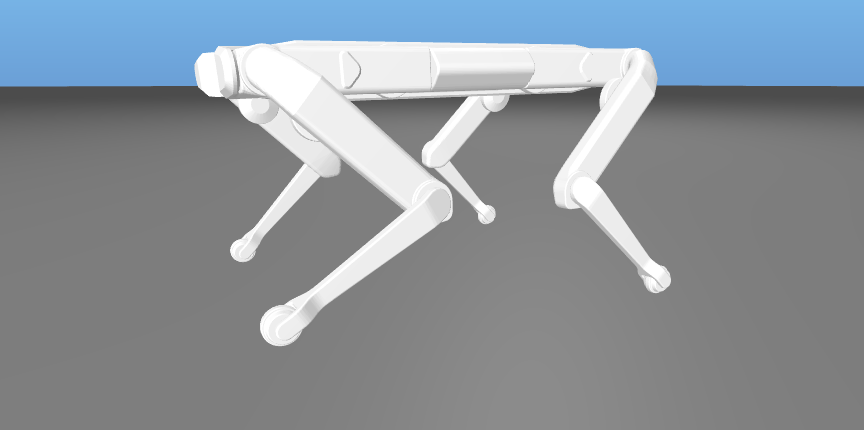}
    \includegraphics[width=0.15 \textwidth]{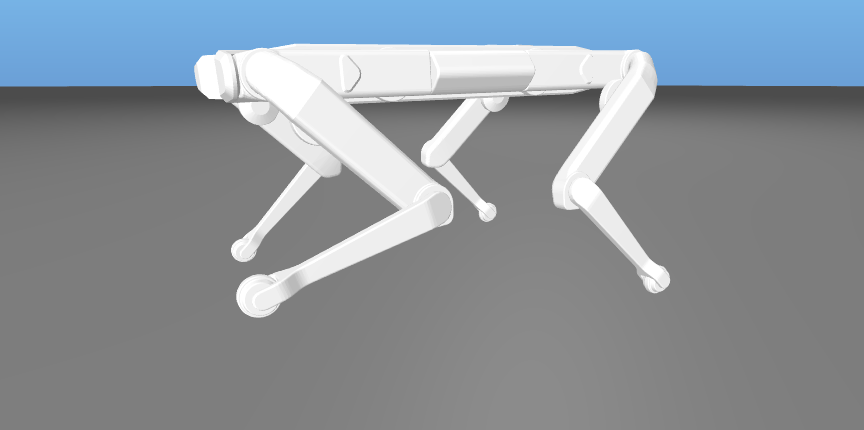}
    \includegraphics[width=0.15 \textwidth]{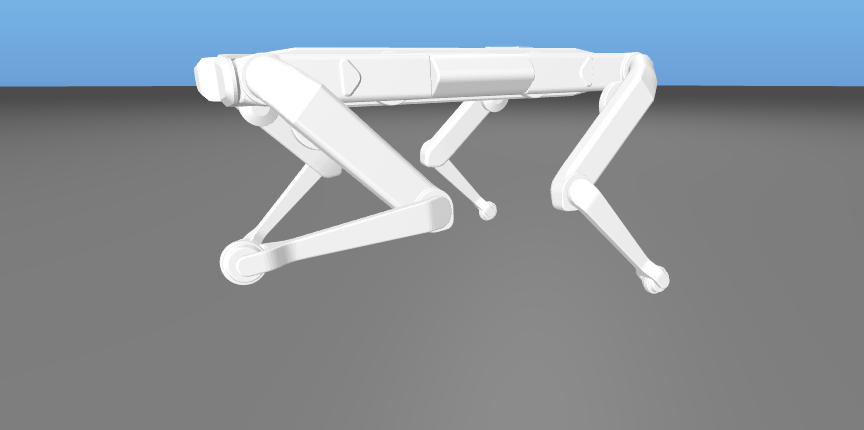}
    \includegraphics[width=0.15 \textwidth]{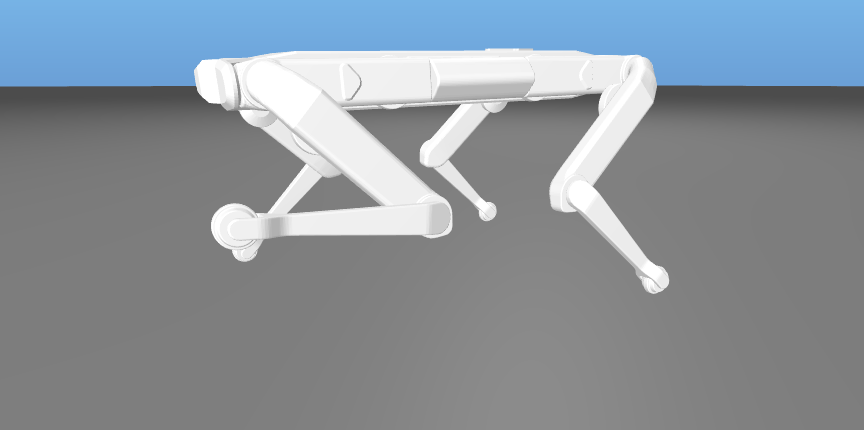}
    \caption{Randomized-DDP leverages the smooth dynamics to precisely schedule movements requiring to break contacts.}
    \label{fig:picture_solo}
\end{figure*}

\begin{figure}[t]
    \centering
    \includegraphics[width=0.45 \linewidth]{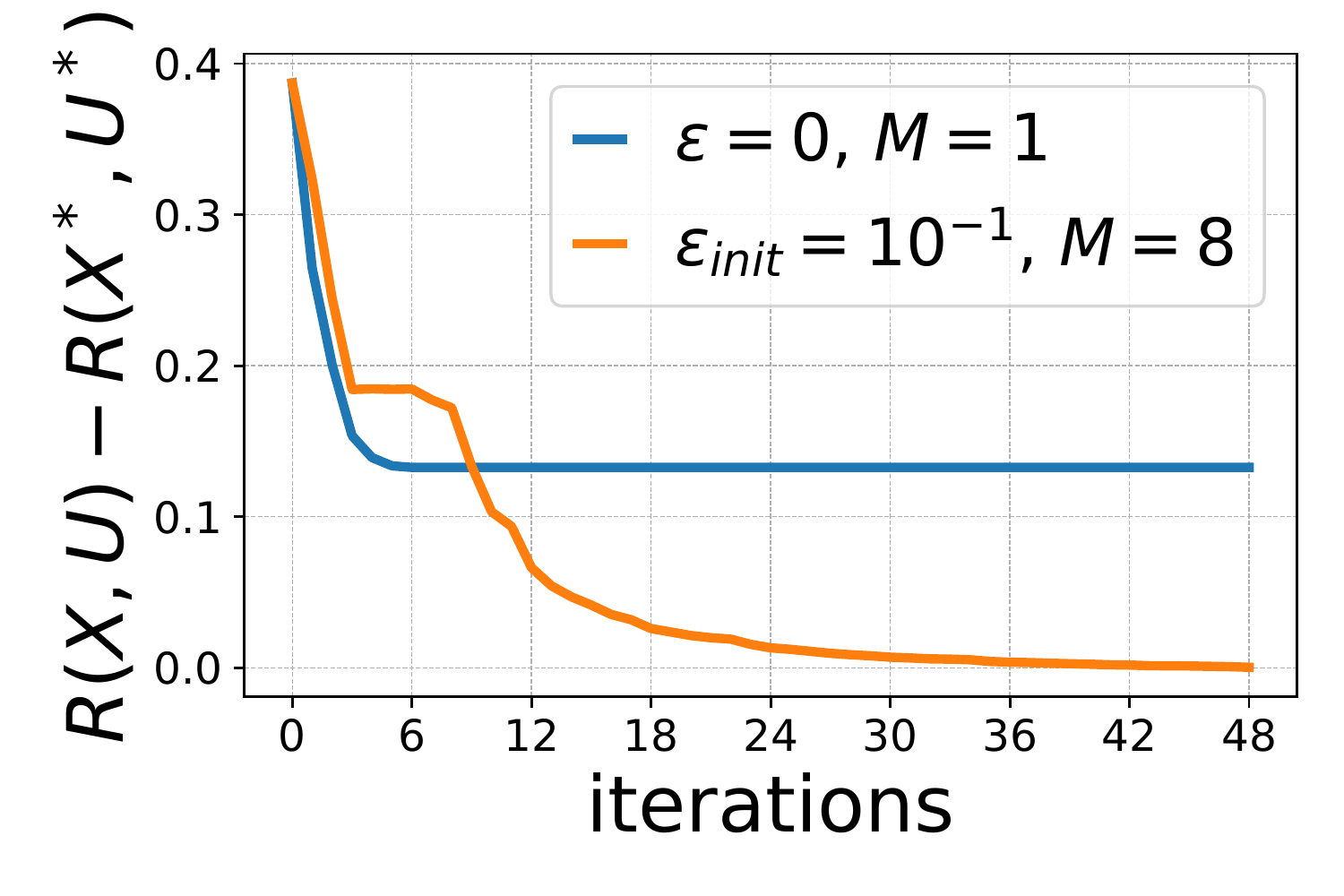}
    \includegraphics[width=0.45 \linewidth]{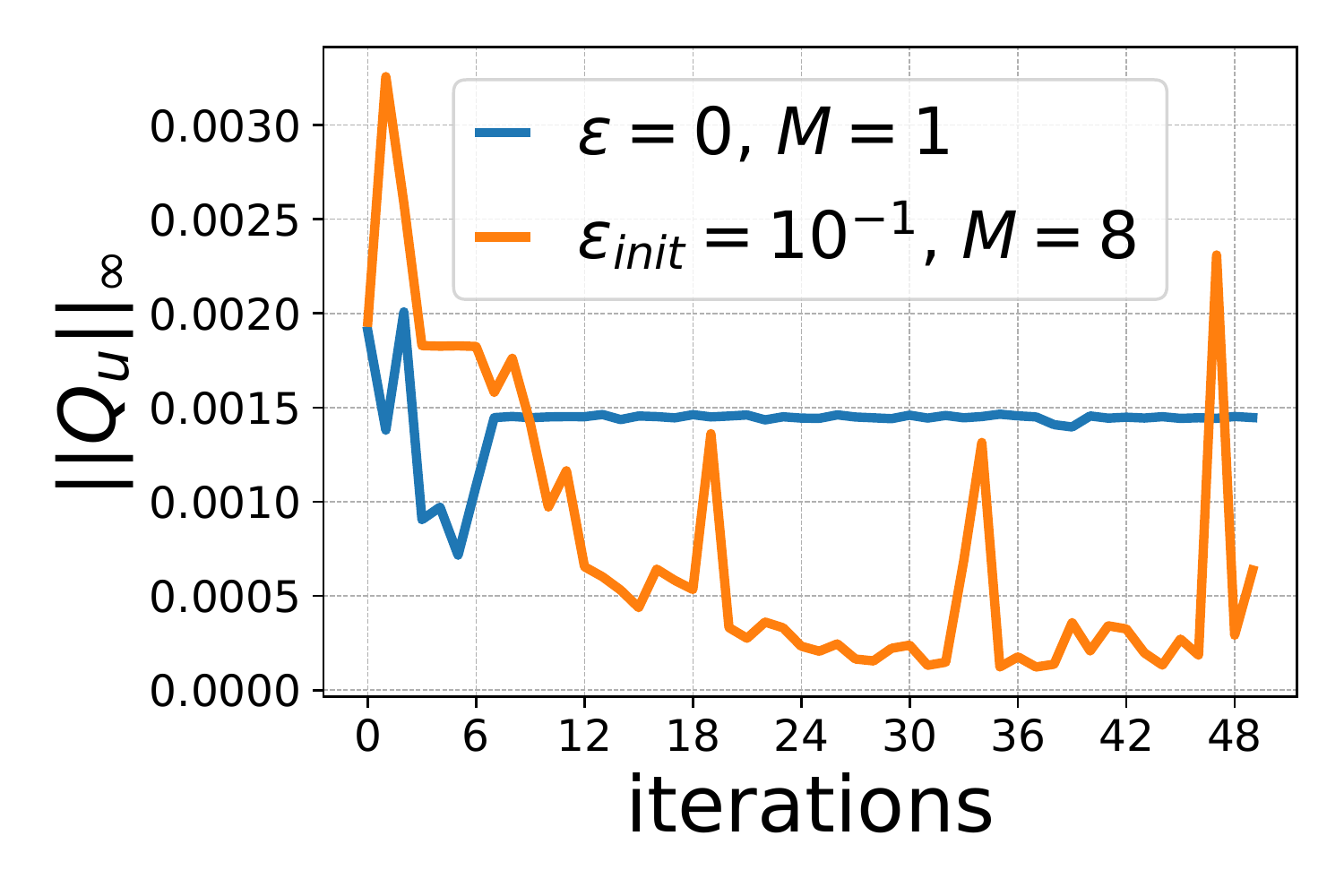}
    \caption{Randomized DDP makes it possible to achieve complex tasks requiring breaking contacts with the floor on Solo robot such as lifting a leg (see Fig.~\ref{fig:picture_solo})}
    \label{fig:solo_lift_foot}
\end{figure}
Finally, we apply our algorithm to solve a task on the Solo robot~\cite{grimminger2020open}, an 18-DoF robotics system, with frictional contacts.
Here, the goal is to reach a final target pose (lifting of the tip of one leg while keeping the three others on the ground), which requires breaking some initial existing contacts, as illustrated in Fig.~\ref{fig:picture_solo}.
Randomized smoothing coupled with Differential Dynamic Programming (R-DDP) allows to solve this task while classical DDP algorithm is not able to precisely apprehend contacts because of non-informative gradients of the dynamics (Fig.~\ref{fig:solo_lift_foot}).
Typically, the control obtained from DDP will only approach this pose while maintaining the feet on the ground which results in Solo bending its leg instead of lifting it.

%% file: sections/5_discussion/discussion.tex
Gradients from differentiable physical processes should not be thought about separately from the actual problem and the associated optimization algorithm.
It is often the case that exact gradients estimated from analytical expressions or automatic differentiation are likely to provide insufficient information for classical optimization to get satisfying solutions.
Randomized smoothing proposes to capture additional information by using a noise distribution to evaluate gradients in the neighborhood of the current optimization stage.
Doing so, it moves from a deterministic to a stochastic setting which may hinder performance of optimization algorithms (convergence and reproducibility) and thus requires some adjustments.
We propose an approach which decreases the noise adaptively to improve robustness.
However, detecting when to increase it is left as future work.
Alternatively, analysing the respective stability properties of the zero-th and first order estimators represents another promising research direction \cite{suh2022differentiable}.